\documentclass[oneside, 12pt, a4paper]{article}

\usepackage[authoryear]{natbib}

\usepackage{geometry}

\makeatletter
\renewcommand\@biblabel[1]{\textbullet}
\makeatother
\newcommand{\appropto}{\mathrel{\vcenter{
			\offinterlineskip\halign{\hfil$##$\cr
				\propto\cr\noalign{\kern2pt}\sim\cr\noalign{\kern-2pt}}}}}

\usepackage[utf8]{inputenc}
\usepackage{verbatim}

\usepackage{todonotes}
\usepackage{amsmath}
\usepackage{amsfonts}
\usepackage{amssymb}
\usepackage{amsthm}
\usepackage{mathtools}
\usepackage{enumerate}
\usepackage{fullpage}
\usepackage{accents}
\usepackage{makeidx}
\usepackage{pdfpages}
\usepackage[parfill]{parskip}
\usepackage{color}
\usepackage{titlesec}
\usepackage{float}
\usepackage{colortbl}
\usepackage{listings}
\usepackage{fancyvrb}
\usepackage{graphicx}
\usepackage{caption}
\titleformat*{\section}{\large\bfseries\centering}
\titleformat*{\subsection}{\bfseries}

\usepackage{bbm}
\usepackage{framed}
\def\sJ{{\mathcal J}}                            
\def\sl{{\ell}}                                  

\def\sT{{\mathcal T}}

\def\sO{{\mathcal O}}

\def\sA{{\mathcal A}}

\def\sU{{\mathcal U}}

\def\vectorfontone{\bf}
\def\vectorfonttwo{\boldsymbol}
\def\vx{{\vectorfontone x}}                      
\def\vy{{\vectorfontone y}}                      
\def\vz{{\vectorfontone z}}                      %

\def\vtheta{{\vectorfonttwo \theta}}             
\def\vvartheta{{\vectorfonttwo \vartheta}}       %
\def\vmu{{\vectorfonttwo \mu}}                   
\def\vnu{{\vectorfonttwo \nu}}                   %
\def\vpi{{\vectorfonttwo \pi}}                   %
\def\vrho{{\vectorfonttwo \rho}}                 %
\def\vsigma{{\vectorfonttwo \sigma}}             %
\def\matrixfontone{\bf}
\def\matrixfonttwo{\boldsymbol}
\def\mA{{\matrixfontone A}}                      %
\def\mB{{\matrixfontone B}}                      %
\def\mG{{\matrixfontone G}}                      
\def\mS{{\matrixfontone S}}                      %
\def\mX{{\matrixfontone X}}                      
\def\mY{{\matrixfontone Y}}                      %

\def\vY{{\matrixfontone Y}}

\def\mGamma{{\matrixfonttwo \Gamma}}             %
\def\mSigma{{\matrixfonttwo \Sigma}}             %
\def\bE{{\mathbb E}}                             
\def\bR{{\mathbb R}}                             
\def\bI{{\mathbb I}}                             

\def\ds{\displaystyle}

\theoremstyle{plain}

\theoremstyle{definition}

\newcommand\numberthis{\addtocounter{equation}{1}\tag{\theequation}}

\definecolor{light-gray}{gray}{0.95}

\newcommand{\code}[1]{\colorbox{light-gray}{\texttt{#1}}}

\begin{document}

\begin{center}
	{\Large Diagonal Discriminant Analysis with Feature Selection for High Dimensional Data}\\
	\vspace{2mm}
	{\large Sarah \textsc{E. Romanes}${}^{(1)}$, John \textsc{T. Ormerod}${}^{(1,2)}$, and Jean \textsc{Y.H. Yang}${}^{(1,3)}$}
\end{center}

\medskip	
\medskip
\centerline{\it ${}^{(1)}$ School of Mathematics and Statistics, University of Sydney, Sydney 2006, Australia}
\medskip
\centerline{\it ${}^{(2)}$ARC Centre of Excellence for Mathematical \& Statistical Frontiers,}
\smallskip 
\centerline{\it The University of Melbourne, Parkville VIC 3010, Australia}
\medskip
\centerline{\it ${}^{(3)}$ The Judith and David Coffey Life Lab, Charles Perkins Centre,}
	\smallskip  
\centerline{\it University of Sydney, Sydney 2006, Australia}

\medskip

\begin{abstract}
	We introduce a new method of performing high dimensional discriminant analysis, which we call multiDA. We  achieve this by constructing a hybrid model that seamlessly integrates a multiclass diagonal discriminant analysis model and feature selection components. Our feature selection component naturally simplifies to weights which are simple functions of likelihood ratio statistics allowing natural comparisons with traditional hypothesis testing methods. We provide heuristic arguments suggesting desirable asymptotic properties of our algorithm with regards to feature selection.  We compare our method with several other approaches, showing marked improvements in regard to prediction accuracy, interpretability of chosen features, and algorithm run time. We demonstrate such strengths of our model by showing strong
	classification performance on publicly available high dimensional datasets, as well as through multiple simulation studies. We make an R package available implementing our approach.
\end{abstract}

\noindent%

\vfill

\vspace{1cm}

\section{Introduction}\label{Intro}

	Classification problems involving high dimensional data are extensive in many fields such as finance, marketing, and bioinformatics. Unique challenges with high dimensional datasets are numerous and well known, with many classifiers built under traditional low dimensional frameworks simply unable to be applied to such high dimensional data.  Discriminant Analysis (DA)  is one such example \citep{fisher1936}. DA classifiers work by assuming the distribution of the features is strictly Gaussian at the class level, and assign a particular point to the class label which minimises the Mahalanobis (for linear discriminant analysis (LDA)) distance between that point and the mean of the multivariate normal  corresponding to such class.  Although extraordinary simple and easy to use in low dimensional settings, DA is well known to be unusable in high dimensions due to the maximum likelihood estimate of the 
	corresponding covariance matrix being singular when the number of features is greater than that of the observations.

	 One alternative to DA to permit use in high dimensions is known diagonal DA (also known as n\"aive Bayes) \citep{Freedman1989,dudoit2002,bickel2004}, which makes the simplifying assumption that all features are independent. As such the resultant covariance matrix is diagonal - circumventing aforementioned issues associated with singular covariance matrix estimates. While this assumption is strong, \cite{bickel2004} show that the classification error for diagonal DA can still work well in high dimensional settings
	 even when there is dependence between predictors.    Another significant challenge in high dimensional frameworks is associated with feature selection, not only to improve accuracy of classifiers, but also to facilitate interpretable applications of results. As such, one obvious drawback of diagonal DA is that it depends on all available features, and as such is not informative as to which features are important.
	
	As such, many variants of DA have been developed in order to integrate feature selection, many resulting in classifiers that involve a sparse linear combination of the features. \citet{Pang2009} improved diagonal DA through shrinkage and regularisation of variances, whilst others have achieved sparsity through regularising the log-likelihood for the normal model or the optimal scoring problem, using the $L_1$ penalty \citep[see][]{Ref:tibshirani2003, LENG2008, clemmensen2011}. \citet{Ref:penLDA} approached the problem through Fisher's discrimination framework - reshaping Fisher's discriminant problem as a biconvex problem that can be optimised using a single iterative algorithm. Whilst classifiers such as these offer significant improvements to the original diagonal DA algorithm, their successful implementation in many cases is dependent upon tuning parameters. These parameters, which in order to tune properly, requires either knowledge of the true sparsity of model parameters, or more commonly needs to be tuned through cross validation - subsequently increasing the run time of these procedures. Furthermore, many of these methods assume equality of variances between the classes, and as such do not perform well when this assumption is not met.
	
	An alternative (and possibly, the simplest) approach to feature selection for diagonal DA classifiers is to recast the problem as one focused on \textit{multiple hypothesis testing}, with such testing performed across all features and information about their significance used to drive classification. Not only does this approach have advantages in its conceptual simplicity, it also generates intuitive insights into features that drive prediction. 
	Most importantly, however, as the number of tests and inferences made across the features increases, so too does the probability of generating erroneous significance results by chance alone, especially in a high dimensional feature space. Many procedures have been developed to control Type I errors, with the False Discovery Rate (the expected proportion of Type I errors among the rejected hypotheses) and  the Family Wise Error Rate (the probability of at least one Type I error)  being the most widely used control methods \citep[see][]{Bonferroni36, Holm1979, hochberg1995}.
	Alternatively, one may control the number of significant features within the multiple hypothesis testing framework by penalising the likelihood ratio test statistic between hypothesis being compared. Not only does this control the number of selected features, this methodology allows for one to weight the features meaningfully for prediction, rather than weighting all significant features equally.
	
	Many other classifiers exist all together outside of the realm of DA to tackle high dimensional problems. These include, but are not limited to: Random Forest \citep{Ref:Breiman2001}, Support Vector Machines (SVM) \citep{Ref:Cortes1995}, K-Nearest Neighbours (KNN) \citep{Ref:KNNPaper}, and logistic regression using LASSO regularisation \citep{Ref:lasso1996}. Although some of these classifiers have demonstrated considerable predictive performance, many lack insight into features driving their predictive processes. For example, an ensemble learner model built using Random Forest with even moderate tree depth of 5, say, may already have hundreds of nodes, and whilst such a model may have strong predictive abilities, using it as a model to explain driving features is almost impossible. The importance of interpretable classifiers is not to be understated, with in many fields such as clinical practice, complex machine learning approaches often cannot be used as their predictions are difficult to interpret, and hence are not actionable \citep{Lundberg206540}.
	
	In this paper we propose a new high dimensional classifier, utilising the desirable generative properties of diagonal DA and improved through treating feature selection as a multiple hypothesis testing problem, resulting not only in a fast and effective classifier, but also providing intuitive and interpretable insights into features driving prediction. We call this method multiDA. Furthermore, we are able to relax the assumption between equality of group variances. We achieve this by:
	\begin{enumerate}
		\item Setting up a probabilistic frame work to consider an exhaustive set of class partitions for each variable;
		
		\item Utilising latent variables  in order to facilitate inference for determining discriminative features, with a penalised likelihood ratio test statistic estimated through maximum likelihood estimates providing the foundation of our multiple hypothesis testing approach to feature selection; and
		
		\item Using estimated weights for each variable to provide predictions of class labels effectively.
	\end{enumerate} 

	Our multiple hypothesis testing paradigm is different to those methods which seek to
	control Type I errors, e.g., controlling the False Discovery Rate or Family Wise Error Rate.
	Instead we attempt to achieve Chernoff consistency where we aim to have asymptotically
	zero type I and type II errors in the sample size \citep[see Chapter 2 of][for a formal definition]{Shao2003}.
	We will also provide heuristic asymptotic arguments of the Chernoff consistency of our multiple hypothesis framework utilised for feature selection (see Appendix A). Our latent variable approach is similar manner to the Expectation Maximization Variable Section (EMVS) approach of \cite{Rockova2014} who uses latent variables
	in the context of variable selection in linear models.
	
	Lastly, we develop an efficient \code{R} package implementing our approach we name \code{multiDA}.  This is available from the following web address:
		\begin{center}
			\code{http://www.maths.usyd.edu.au/u/jormerod/} 
		\end{center}
		
	The outline of this paper is as follows. In Section \ref{hypothesis_test}, we will discuss the role of hypothesis testing in selecting discriminative features. In Section \ref{model} we introduce our multiDA model, in Section \ref{model_Est} discuss the model estimation procedure, and in Section \ref{theo} we discuss some theoretical considerations for our model. 
	Further, in Section \ref{numerical} we demonstrate the performance of our model under simulation and also with publicly available datasets. 

\section{Identifying discriminative features}\label{hypothesis_test}

	 Consider data which may be summarized by an $n \times p$ matrix $\mX$ with elements $x_{ij}$, where $x_{ij}$ denotes the value for sample $i$ of feature $j$. For $K$ classes, class labels the $n$-vector $\vy$ whose elements are integers ranging from $1$ to $K$, with $n_k$ denoting the number of observations belonging to class $k$. Further, let $\mY$ be a matrix of dimension $n \times K$, with $Y_{ik}$ a binary variable indicating whether observation $i$ is in class $k$, and $\vY_i = (Y_{i1}, \dots, Y_{iK})^T$. We assume that class labels $\vy$ are assigned to all observations, and combined with the matrix $\mX$ form what is referred to in machine learning literature as the \textit{training} dataset.
	 
	Recall that for diagonal DA, we assume the features are conditionally independent given the class label. As such, the likelihood (assuming equality of variances between the groups being compared) is simply:
\begin{equation}\label{eq:dlda}
p(\mX, \mY;\vtheta)=\prod_{i=1}^n\left[ \prod_{j=1}^p \prod_{k=1}^K \phi(x_{ij}; \mu_{jk}, \sigma^2_{jk})^{Y_{ik}} \times \prod_{k=1}^K \pi_k^{Y_{ik}} \right],
\end{equation}

\noindent where $\phi(x_{ij}, \mu_{jk}, \sigma^2_{jk})$ represents a Gaussian distribution with $\mu_{jk}$ being the mean of feature $j$ in observations of class $k$, and $\sigma^2_{jk}$ is its variance. Further, $\pi_k$ represents the class prior - that is, the probability of an observation being in class $k$, and 
$\vtheta = (\vmu, \vsigma^2, \vpi)$
with $\vmu = (\mu_{jk})_{1\le j\le p, 1\le k\le K}$, 
$\vsigma^2 = (\sigma_{jk}^2)_{1\le j\le p, 1\le k\le K}$, and 
$\vpi = (\pi_1,\ldots,\pi_K)^T$.

We will later model a latent variable 
	extension to handle (A) multiple hypothesis testing within each variable, (B)
	multiple hypothesis testing across variables, (C) provide theory showing how
	Chernoff consistency can be maintained, and (D) how predictions can be made. 
	However, before we do this we motivate our approach by describing how 
	hypothesis testing can be used to identify discriminative features, first
	when there are two classes, then three classes, and then the general case.
 
\subsection{Two classes} 

\begin{figure}[h]
		\begin{center}
		\includegraphics[width=0.7\textwidth]{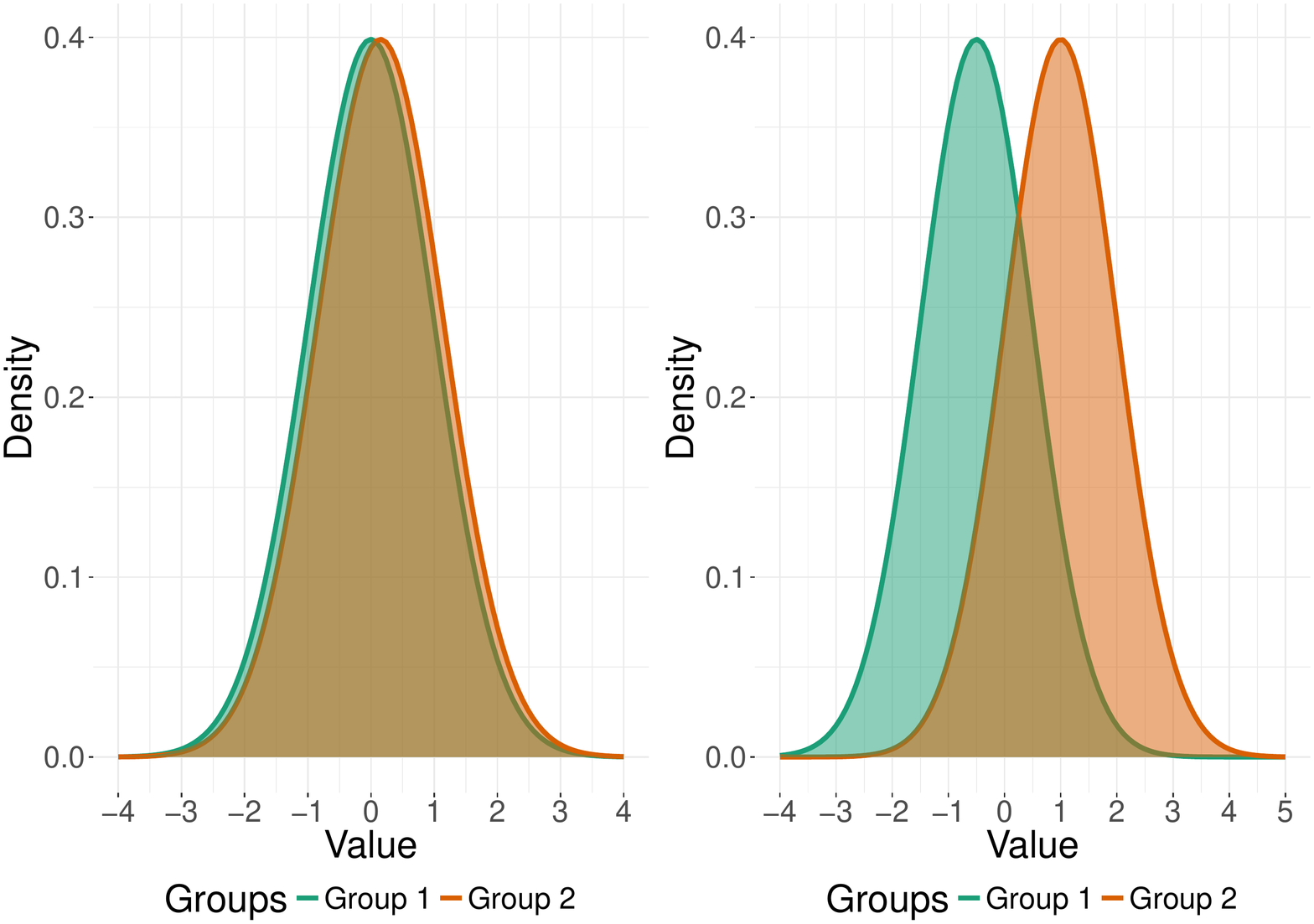}
	\end{center}
		\caption{\textbf{Left:} An example of a non discriminative feature (significant overlap between class densities). \textbf{Right:} An example of a discriminative feature.}
		\label{fig:TwoClassComparison}
	\end{figure}

	Consider a single feature $j$, and suppose that $K=2$. We will assume that $x_{ij}$ are normally distributed, given class labels.
	For simplicity, we will assume that the variances are the same for each class (later we will relax this assumption). In the most basic form, one can set up the hypothesis testing framework as a simple two sample $t$-test, that is to test the claim that the group means are equal for a particular feature $j$, vs the alternative that they are not. The corresponding hypotheses are
	$$
	H_{j0} \colon \mu_{j1} = \mu_{j2}
	\qquad \mbox{versus}  \qquad 
	H_{j1} \colon \mu_{j1} \ne \mu_{j2}
	$$
	
	where $\mu_{j1}$ and $\mu_{j2}$ are the class means for class 1 and class 2 respectively, as illustrated in Figure \ref{fig:TwoClassComparison}.
	
	It is clear that when comparing two classes, that there is only one way to partition the groupings to determine discriminative features, that is, either the group means are equal (non discriminative), or they are not (discriminative). However, when the number of classes increases beyond the binary case, the number of groupings is more nuanced. 
	
	\subsection{Three classes} 
	
	Assuming equal variances and $K= 3$ there are four different ways that
	a feature can be discriminative and 
	one way a feature can be non-discriminative. The  non-discriminative
	case corresponds to the case where the mean for each class is the same.
	The discriminative cases are
	\begin{itemize}
		\item One of the class means is different from the other two class means (there are 3 ways that this can happen); and
		
		\item Each of the class means is different from each other (there is one way that this can happen).
	\end{itemize}

	We will capture the ways that the classes can be grouped together via a $K \times M$ matrix $\mS$ where $M$ is the number of hypothesised models. For the above situation the corresponding $\mS$ matrix is of the form
	\begin{equation}\label{eq:Smatrix}
	\mS =\begin{bmatrix}{}
	1 &   1 &   1 &   2 &   1 \\ 
	1 &   2 &   1 &   1 &   2 \\ 
	1 &   1 &   2 &   1 &   3 \\ 
	\end{bmatrix}.
	\end{equation}
	

	Consider the notation for model parameters for the case $K=3$ for each of the $M=5$ hypotheses ($m=1,\ldots,M$) for variable $j$ we have a
	single variance $\sigma_{jm}^2$ for each hypothesis.  Define the vector $\mG$ as the number of partitions in each column of the set partition matrix $\mS$. In the case
	of (\ref{eq:Smatrix}) we have $\mG = (1,2,2,2,3)^T$. Next we let 
	$\mu_{jmg}$ denote the mean corresponding to variable $j$, for the $m$th hypothesis,
	in the $g$th group.
	In order to breakdown the parametrisation of the class means
	consider the cases:
	\begin{itemize}
	\item When $m=1$, the 1st column of $\mS$, i.e., $\mS_1 = [1,1,1]^T$ we have one group (so that $G_1=1$)
	and $g=1$ and one mean $\mu_{j11}$.
	This corresponds to the null case where there is no differences in the distribution between classes for variable $j$, i.e., variable $j$ is non-discriminative. The corresponding hypothesis is
	$$
	H_{j1} \colon x_{ij} \sim \phi( \; \cdot \; ; \mu_{j11},\sigma_{j1}^2).
	$$
	
	\item When $m=2$, the 2nd column of $\mS$, i.e., $\mS_2 = [1,1,2]^T$ we have two groups (so that $G_2=2$)
	and $g \in \{ 1,2 \}$ with two means mean $\mu_{j21}$ and $\mu_{j22}$. 
	The corresponding hypothesis is
	$$
	H_{j2} \colon x_{ij} \sim \phi( \; \cdot \; ; \mu_{j21},\sigma_{j2}^2)^{\bI(y_{i} \in\{1,2\})} 
	\phi( \; \cdot \; ; \mu_{j22},\sigma_{j2}^2)^{\bI(y_{i}=3)}.
	$$
	
	\noindent Similarly for $m=3$ and $m=4$.
	
	\item When $m=5$, the 5th column of $\mS$, i.e., $\mS_5 = [1,2,3]^T$ each class has a different mean
	(so that $G_5 = 3$)  and $g \in \{ 1, 2, 3 \}$ with three means mean $\mu_{j51}$, $\mu_{j52}$
	and $\mu_{j53}$. The corresponding hypothesis is
	$$
	H_{j5} \colon x_{ij} \sim 
	\phi( \; \cdot \; ; \mu_{j51},\sigma_{j5}^2)^{\bI(y_{i}=1)} 
	\phi( \; \cdot \; ; \mu_{j52},\sigma_{j5}^2)^{\bI(y_{i}=2)} 
	\phi( \; \cdot \; ; \mu_{j23},\sigma_{j5}^2)^{\bI(y_{i}=3)}.
	$$
	
\end{itemize}

	\subsection{More than three classes} 
	
	For a particular feature $j$, it is clear that for $K>2$ there are multiple alternate hypothesis to consider when testing against the null and alternate distributions, with there being $B_K$ (the Bell number of order $K$) ways one could partition $K$ objects \citep[for R implementation, see][]{hankinParts}.   Let $\gamma_{jm}$ be a binary variable, indicating whether feature $j$ belongs in partition set $m$, where $m=1, \ldots,M$. We note that the first column of $\mS$ always corresponds to the \textit{null} case (i.e., no differences between the $K$ classes for feature $j$), with the columns increasing in degrees of freedom. We also note that all of the columns are nested within the last column, corresponding to the case in which all the classes are different for a particular feature $j$. Further, define $\vnu$ as a vector describing the degrees of freedom of each partition, relative to the null. 
		For example, in (\ref{eq:Smatrix}) we have $K=3$, $M=B_3 = 5$,  and $\vnu = (0,1,1,1,2)$.

	In general we will use the following notation for model parameters for an arbitrary partition matrix $\mS$. The $m=1,\ldots,M$ hypotheses for variable $j$ are of the form 
	$$
	H_{j m} \colon x_{ij} \stackrel{\mbox{\scriptsize iid}}{\sim} p_{jm}( \;\cdot \;, \mY_i;\vtheta_{jm})
	$$
	
	\noindent where
	\begin{equation}\label{eq:ldaHypotheses}
	p_{jm}(x_{ij} | \mY_i;\vtheta_{jm}) =  
	\prod_{g=1}^{G_m} \phi(x_{ij}; \mu_{jmg}, \sigma^2_{jm})^{\mathbb{I} (\mY_i^T\mS_m = g) }
	\end{equation}
	
	\noindent and $\vtheta_{jm}$ is a vector consisting of all of the 
	$\mu_{jmg}$'s and $\sigma^2_{jm}$'s. The above uses the fact that $\mY_i^T\mS_m$ can
	be used to determine which of the $G_m$ means should be used for sample $i$ and
	model $m$.
	
\subsection{Restricting the number of hypotheses} 

	Note that we can specify many alternate options for $\mS$, as $B_K$ can grow \textit{very} quickly. One popular option for multi-class classification algorithms is to consider the \textit{one vs. rest} approach, in which the multi-class problem reduces down to multiple ``binary'' comparisons by restricting $\mS$ by only considering partitions such that $\max_{1\le m\le M}(G_m) = 2$. In that case, $\mS$ is reduced, with for the previous example, $M = 4 = K + 1$, $\mG=(1,2,2,2)$, and $\vnu = (0,1,1,1)$.
	\begin{align*}
	\mS =\begin{bmatrix}{}
	1 &   1 &   1 &   2  \\ 
	1 &   2 &   1 &   1  \\ 
	1 &   1 &   2 &   1  \\ 
	\end{bmatrix}.
	\end{align*}
	This is reduction in size is particularly useful when $K$ is large. For example, suppose $K=15$. For the full set partition matrix, $M=B_{15}= 1, 382,958,545$, where as for the one vs. rest case, $M=15+1=16$, heavily reducing computational time throughout our resultant algorithm.

	Another possible configuration for $\mS$ is to only consider groupings which are \textit{ordinal}. This may be particularly useful for datasets in which the classes are ordinal in nature, say, classifying stages of cancer development. An ordinal configuration of $\mS$ allows some reduction in computational time, whilst leveraging useful information about the classes. In this case, $\mS$ reduces to:
	\begin{align*}
	\mS =\begin{bmatrix}{}
	1 &    1 &   2 &   1 \\ 
	1 &    1 &   1 &   2 \\ 
	1 &    2 &   1 &   3 \\ 
	\end{bmatrix}.
	\end{align*}
	
	If the classes had an ordinal structure, such as stages of cancer, we might assume that the mean might change
	from stages 1 to stages 2, but not back to the mean of stages 1 for class 3.
	This removes the case $[1,2,1]$ from (\ref{eq:Smatrix}) since under this model
	class 1 and class 3 are grouped together with class 2 having a different mean.

	\subsection{Heterogeneity of group variances}\label{heterogeneity}
	
\begin{figure}[h]
		\begin{center}
			\includegraphics[width=0.75\textwidth]{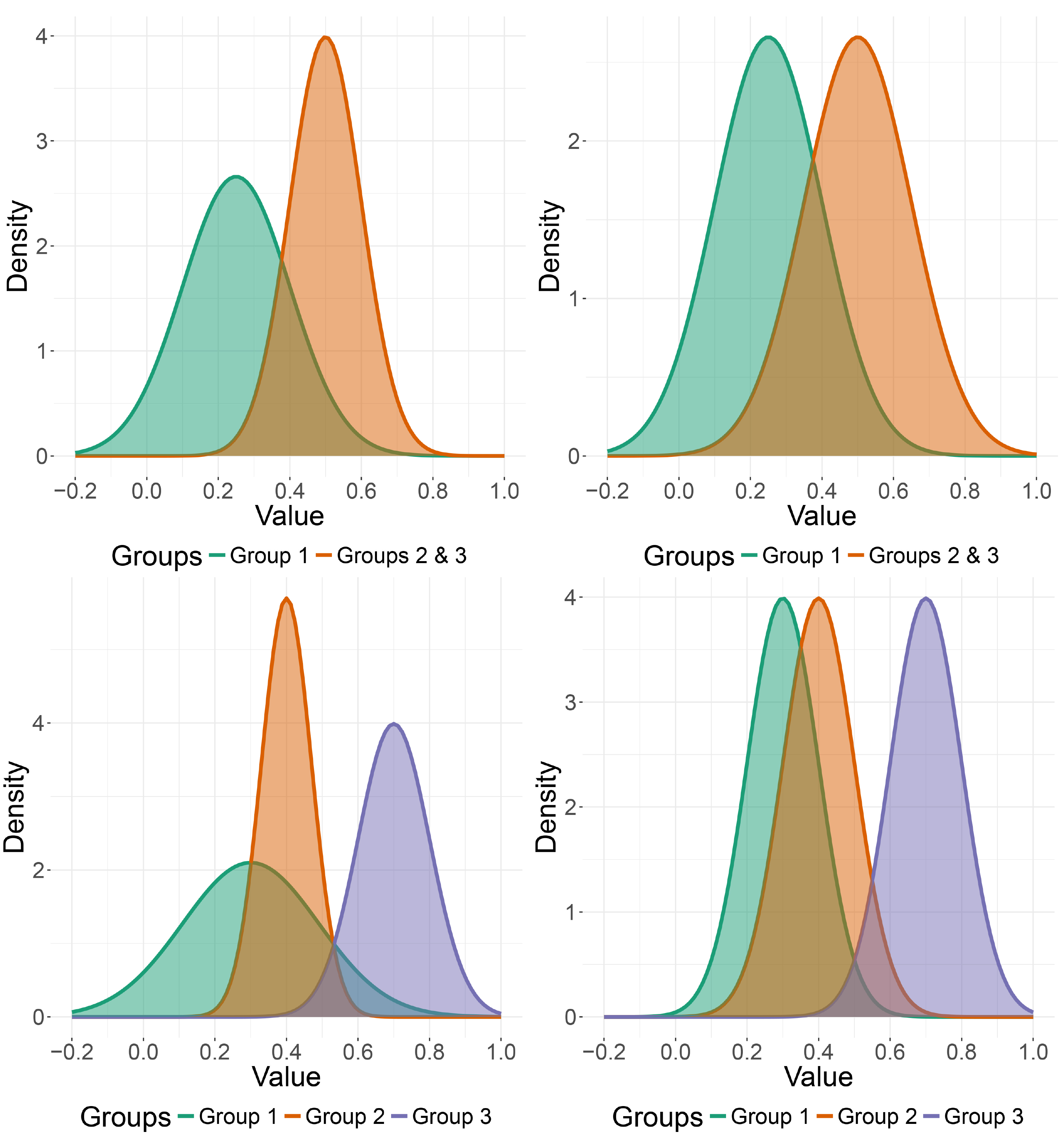}
		\end{center}
		\caption{\textbf{Above:} Example of one-vs-rest separation for three cases with and without equality of variance assumption.
			\textbf{Below:} Example of complete separation for three cases with and without equality of variance assumption.} 
		\label{fig:Eg2}
	\end{figure}

	
	In many DA papers published previously, the group specific variances are assumed to be equal. 
	This assumption leads what is known as a Linear Discriminant Analysis (LDA) classifier since the resultant classification boundary is linear in the predictor variables. We will refer to our implementation of LDA as \textit{multiLDA}. 
	However, if we generalise and do not make such an assumption, the resultant classification boundary  is instead quadratic  and our classifier is in the form of what is known as a Quadratic Discriminant Analysis (QDA) classifier. The QDA case leads to $\sum_{m=1}^{M} G_m$  variance estimates required for each feature $j$ (as opposed to $M$ estimates required under the multiLDA framework). The model
	for $x_{ij}$ when assuming unequal variances is:
	$$
	\begin{array}{rl} 
p_{jm}(x_{ij} | \mY_i;\vtheta_{jm})  
	& \ds = \prod_{g=1}^{G_m} \phi(x_{ij}; \mu_{jmg}, \sigma^2_{jmg})^{\mathbb{I} (\mY_i ^T \mS_m = g) } 
	\end{array} 
	$$
	
	\noindent 
	and will be referred to as \textit{multiQDA}. 
	Figure \ref{fig:Eg2} illustrates some simple hypotheses we consider in this paper, including cases in which variances are not assumed to be equal.

\section{The multiDA model}\label{model}


 We now describe our 
 multiDA model which is capable of handling not only multiple hypotheses
across variables, but also multiple hypotheses
within variables.
	When adapting the diagonal DA model to account for multiple hypothesis testing, the representation of the likelihood is more complicated than diagonal DA described previously. We now   take the product over the number of hypothesis to be tested, and account for the different number of normal distributions in each hypothesis test. We account for this, and now consider fitting a model of the form:
	\begin{equation}\label{eq:multiDa}
		p(\mX; \vtheta|\mY,\mGamma) = \prod_{j=1}^{p}\prod_{m=1}^{M}[p_{jm}(\vx_j| \mY_i;\vtheta_{jm})]^{\gamma_{jm}}
	\end{equation}
	
	where 
$$
(Y_{i1}, \dots, Y_{iK})|\vpi \sim \text{Multinomial}(1;\pi_1, \dots, \pi_K),
$$

\noindent with $0\leq \pi_k \leq 1$ and $\sum_{k=1}^{K} \pi_k =1$ being the class probabilities
to be estimated,
	the $\gamma_{jm} \in \{0,1\}$ are latent variables satisfying 
	$$
	(\gamma_{j1}, \dots, \gamma_{jM})|\vrho_j \sim \text{Multinomial}(1;\rho_1, \dots, \rho_M),
	$$
	
	\noindent with $0\leq \rho_m \leq 1$ and $\sum_{m=1}^{M} \rho_m =1$,
	and $p_{jm}(\vx_j, \vy;\vtheta_{jm})$ is a model likelihood for variable $j$ and model $m$ with
	corresponding parameters $\vtheta_{jm}$. The $\gamma_{jm}$ play the role of selecting which of the
	models $m \in \{1,\ldots, M\}$ is used for variable $j = 1,\ldots, p$.
	Here $\vtheta = (\vtheta_{jm})_{1\le j\le p, 1\le m\le M}$. 
	The $\rho_m$ are prior probabilities for the $m$th hypothesis, in which we are implicitly assuming that for each $m$ the same hypothesis is being tested for each $j=1,\dots, p$. We will treat
	the prior hyperparameters $\rho_1, \dots, \rho_M$ as tuning parameters set by the user (discussed later).
	We can now fully represent our model as:
	\begin{equation}\label{eq:full}
	p(\mX, \mY, \mGamma; \vvartheta) = 
	\left[ \prod_{i=1}^n \prod_{j=1}^p
	\prod_{m=1}^{M}
	\left\{p_{jm}(\vx_j, \vy;\vtheta_{jm})  \right\}^{\gamma_{jm}}
	\right] 
	\times \left[ \prod_{k=1}^K \pi_k^{Y_{ik}} \right] 
	\times \left[ \prod_{j=1}^{p}\prod_{m=1}^{M} \rho_m ^{\gamma_{jm}} \right] 
	\end{equation}
	
	\noindent where $\vvartheta =  (\vtheta,\vpi)$ are likelihood parameters to be estimated.

\section{Model estimation}\label{model_Est}

The first goal of the multiDA is to determine which features partition the data into meaningful groups, that is, estimating posterior probabilities of the $\gamma$'s. To do this, we will fit this model using maximum likelihood over $\vvartheta$. The likelihood is given by
$$
\mathcal{L}(\vvartheta) = \sum_{\mGamma}p(\mX| \mY, \mGamma; \vtheta) p(\mY;\vpi )p(\mGamma; \vrho).
$$

\noindent 
where $\sum_{\mGamma}$ denotes summation over all $M^K$ possibilities of $\mGamma$.

Note that for any fixed value of $\mGamma$, the value of the $\vtheta_{jm}$ which maximises $p(\mX| \mY, \mGamma; \vtheta)$ is the same. Hence, the values of $\vtheta_{jm}$ which maximise $\mathcal{L}(\vvartheta)$ must be these values which are given by
\begin{align*}
\widehat{\vtheta}_{jm} =\arg \max_{\vtheta_{jm}}\{ \; p_{jm}(\vx_j, \mY;\vtheta_{jm}) \; \}
\end{align*}

\noindent where 
$p_{jm}(\vx_j, \mY;\vtheta_{jm}) = \sum_{i=1}^n p_{jm}(x_{ij} | \mY_i;\vtheta_{jm})$.

For the multiLDA case
the explicit log-likelihood given $\mY$ and $\mGamma$ is 
\begin{align*}
\log \mathcal{L} (\vvartheta| \mY, \mGamma) &=\sum_{i=1}^{n} \Bigg[ \sum_{j=1}^{p} \sum_{m=1}^{M} \gamma_{jm} \Bigg\{ 
\sum_{g=1}^{G_m} \sum_{i \in \sA_{mg}}  \log \phi(x_{ij}; \mu_{jmg}; \sigma^2_{jm}) + \log(\rho_v) \Bigg\} \Bigg]\\
& + \sum_{k=1}^{K} n_k\log(\pi_k).
\end{align*}

\noindent where $n_k = \sum_{i=1}^n Y_{ik}$
and $\sA_{mg} = \{ \; i \colon \mY_i^T \mS_m = g \; \}$.
Next, let
$$
n_{mg} = |\sA_{mg}|,
\qquad 
x_{\bullet j mg} = \sum_{i \in \sA_{mg}}  x_{ij}
\qquad \mbox{and} \qquad 
x_{\bullet mg j}^2 = \sum_{i \in \sA_{mg}}  x_{ij}^2.
$$

\noindent Then the MLEs for the $\mu$'s and $\pi$'s may be written as 
$$
\ds \widehat{\mu}_{jmg} = \frac{x_{\bullet jmg}}{n_{mg}}
\qquad \mbox{and} \qquad 
\widehat{\pi}_{k}= \frac{n_k}{n}
$$

\noindent respectively with $j = 1,\ldots, p$ and $k = 1,\ldots, K$. The MLEs for the $\sigma^2$'s
for the multiLDA and multiQDA cases are
$$
\ds \widehat{\sigma}^2_{jm} = \frac{1}{n}  \left[ \sum_{g=1}^{G_m}
\left\{  x_{\bullet jmg}^2 -  \frac{(x_{\bullet jmg})^2}{n_{mg}}
\right\} \right]
\qquad \mbox{and} \qquad 
\ds \widehat{\sigma}^2_{jmg} = \frac{1}{n_{mg}}  \left[  
  x_{\bullet jmg}^2 -  \frac{(x_{\bullet jmg})^2}{n_{mg}}
 \right]
$$

\noindent respectively for each index $j$, $m$ and $g$.

%
%
%
%
%

\subsection{Fitting the latent parameters}\label{latent}

The latent parameters $\gamma_{jm}$ are indicators for each hypothesis test $m$ and for each variable $j$. We will use the posterior probabilities $P(\gamma_{jm} = 1|\vx_j, \mY, \widehat{\vtheta}_{jm})$ to determine which hypothesis is most likely. These posterior
probabilities become variable specific weights which alter the contribution for each
feature when making predictions.
Next we note that 
$$
\begin{array}{l}
\log p(\mX,\mGamma|\mY ;\widehat{\vvartheta}) 
\\		                                 
\ds \qquad  =  \sum_{j=1}^p \sum_{m=1}^M \gamma_{jm}\left\{ \tfrac{1}{2}\lambda_{jm}(\vx_j, \mY)+ \log(\rho_m/\rho_1)\right\} + \log p_{j1}(\vx_j, \mY;\widehat{\vtheta}_{j1}) +  \log(\rho_1),
\end{array}
$$

\noindent 
where $\lambda_{jm}(\vx_j, \mY) 
= 2\log p_{jm}(\vx_j| \mY;\widehat{\vtheta}_{jm}) 
- 2\log p_{j1}(\vx_j| \mY;\widehat{\vtheta}_{j1})$  are likelihood ratio test (LRT) statistics.
 Note that $\lambda_{j1}(\vx_j, \mY) = 0$ for each $j$ which serves as the ``null'' hypothesis for each variable (that the distribution of $x_{ij}$ is the same for each class).
For the multiLDA case, we have
$$
\lambda_{jm}(\vx_j, \mY) 
= n \log(\widehat{\sigma}^2_{j1})
- n \log(\widehat{\sigma}^2_{jm})
$$

\noindent while for multiQDA we have
$$
\lambda_{jm}(\vx_j, \mY) 
= n \log(\widehat{\sigma}^2_{j1})
- \sum_{g=1}^{G_m} n_{gm}\log(\widehat{\sigma}^2_{jmg}),
$$

then
\begin{align}
\widehat{\gamma}_{jm}(\vx_j, \mY) 
= P(\gamma_{jm} = 1|\vx_j, \mY, \widehat{\vtheta}_{jm}) 
= \frac{\exp\left[  
	\tfrac{1}{2}\lambda_{jm}(\vx_j, \mY)
	+ \log(\rho_m/\rho_1)
	\right]}{\ds \sum_{\sl=1}^M 
	\exp\left[  
	\tfrac{1}{2}\lambda_{j\sl}(\vx_j, \mY)
	+ \log(\rho_\sl/\rho_1)
	\right]
}. 
\end{align}

As such, the $\widehat{\gamma}_{jm}$ can be thought of as a function of a penalised likelihood ratio test statistic, with penalisation provided by the value of $\log(\rho_m/\rho_1)$. 
The choice of this penalty trades off type I and type II errors, which we discuss further in Section \ref{theo}.

\subsection{Prediction of new unlabelled data points}\label{prediction}

Let us now consider the problem of predicting the class vector $\mY^* = (Y_1^*,\ldots,Y_{K}^*)^T$
using a new $p$-vector of predictors $\vx^*$. To perform this task we consider the predictive
distribution
\begin{align*}
p(\mY^*|\mX, \mY, \vx; \widehat{\vvartheta}) 
& = \sum_{\mGamma} p(\mY^*|\mX, \mY, \vx, \mGamma; \widehat{\vvartheta}) \label{eq:pred_y} \numberthis.
\end{align*}

\noindent We note that the above sum is computationally intractable. 
Instead of evaluating this sum exactly we note that
$$
\begin{array}{l} 
\ds \log p(\mY^*|\mX, \mY, \vx; \widehat{\vvartheta}) 
\\ 
\ds \qquad  = \log \left[ \sum_{\mGamma} \frac{p(\mGamma|\mX,\mY;\widehat{\vvartheta}) p(\mY^*|\mX, \mY, \vx, \mGamma; \widehat{\vvartheta})}{p(\mGamma|\mX,\mY;\widehat{\vvartheta})}
\right] 
\\
\ds \qquad \ge \sum_{\mGamma}p(\mGamma|\mX,\mY;\widehat{\vvartheta}) \log\left[
 p(\mY^*|\mX, \mY, \vx, \mGamma; \widehat{\vvartheta})
 - \log p(\mGamma|\mX,\mY;\widehat{\vvartheta}) \right]
\\ 
\ds \qquad = \log \underline{p}(\mY^*|\mX, \mY, \vx; \widehat{\vvartheta})
\end{array} 
$$

where we have used Jensen's Inequality in order to approximate \eqref{eq:pred_y}.

Next, define $\vz$ as the cumulative sum of the maximum number of groups in each column of $\mS$, i.e., $G$, such that
$z_l =\sum_{m=1}^l G_{m}$.
Define the allocation matrix $\mA$, of same dimension as $\mS$, whose elements are given by
$
a_{km}=z_{m} - (G_m - S_{km}).
$
This allocation matrix $\mA$ allocates which $\mu_{jmg}$ and $\sigma^2_{jmg}$ components (if multiQDA) belong to each class. For example, for $K=3$ and its respective $\mS$, the resultant  $\mA$ is given by:
\begin{align*}
\mA =\begin{bmatrix}{}
1 &   2 &   4 &   7 &   8 \\ 
1 &   3 &   4 &   6 &   9 \\ 
1 &   2 &   5 &   6 &   10 \\ 
\end{bmatrix}.
\end{align*}

For mulitLDA we have then for a single prediction that $\mY^*$ is a particular class label $k$ with 
approximate probability
\begin{align*}
\underline{p}(\mY^*|\mX, \mY, \vx; \widehat{\vvartheta}) 
& \propto \exp \left[ \sum_{k=1}^{K} Y_{k}^* \left\{  \sum_{j=1}^{p}\sum_{m=1}^{M} \widehat{\gamma}_{jm}  \log \phi(x_{j}^*; \widehat{\mu}_{ja_{km}}; \widehat{\sigma}^2_{jm})  + \widehat{\pi}_k\log(\widehat{\pi}_k) \right\} \right].
\end{align*}

\noindent Let
$$
\eta_k = \sum_{j=1}^{p}\sum_{m=1}^{M} \widehat{\gamma}_{jm}  \log \phi(x_{j}^*; \widehat{\mu}_{ja_{km}}; \widehat{\sigma}^2_{jm})  + \widehat{\pi}_k\log(\widehat{\pi}_k).
$$

\noindent For multiQDA replace $\widehat{\sigma}^2_{jm}$ with $\widehat{\sigma}^2_{jmg}$
in the above two equations for $\underline{p}(\mY^*|\mX, \mY, \vx; \widehat{\vvartheta})$
and $\eta_k$.
Lastly, since $\mY^*$ follows a multinomial distribution we have
$$
\mY^*|\mX, \mY, \vx; \widehat{\vvartheta} \stackrel{\mbox{\scriptsize approx.}}{\sim} \mbox{Multinomial}(1; \widehat{Y}_{1}^*,\ldots, \widehat{Y}_{K}^*)
$$

\noindent where
$\widehat{Y}_{k}^* =  \exp(\eta_{k})/[\sum_{\ell=1}^{K} \exp(\eta_{\ell})]$.

\section{Theoretical considerations}\label{theo}

Now, recall from Section \ref{model_Est} that $\rho_m$ is fixed, with no restrictions on the $\rho_m$ besides that $\sum_{m=1}^{M} \rho_m =1$ and that  $\rho_m \in [0,1]$, $m = 1,\ldots,M$. Choosing these parameters via, say,
some cross-validation procedure would be computationally expensive. Instead we choose the values of these
using heuristic arguments based on asymptotic theory.

\noindent Suppose that we define the hypothesis testing error $E$ as
$$
E
= \sum_{j=1}^p \sum_{m=1}^M | \widehat{\gamma}_{jm}(\vx_j, \mY)  - \gamma_{0jm}|
$$

\noindent where
$\gamma_{0jm} = I( \mbox{$H_{jm}$ is true})$, $1\le j\le p$, $1\le m\le M$.
We will assume that $\gamma_{0jm_j}=1$ for some $m_j\in \{1,\ldots,M\}$
for all $j$.
Note that it can be shown that
$$
\begin{array}{rl}
E 
& \ds =
	2\sum_{m=2}^M \sum_{j \in \sO_m} \widehat{\gamma}_{jm}(\vx_j, \mY)
+ 
	2\sum_{m=1}^M \sum_{j \in \sU_m} \widehat{\gamma}_{jm}(\vx_j, \mY)
\equiv
E_\sO + E_\sU,

\end{array} 
$$

\noindent where $\sO_m$ is the set of overfitting models for the $m$th hypothesis,
 $\sU_m$ is the set of underfitting models for the $m$th hypothesis, and 
the error terms $E_\sO$ and $E_\sU$ correspond to the 
the 
overfitting  and underfitting models respectively.
We can rewrite $E_\sO$ as
$$
\begin{array}{rl}
E_\sO & \ds = 2\sum_{m=2}^M \sum_{j \in \sO_k} \frac{  	\exp\left[ 
	\tfrac{1}{2}\lambda_{jm}(\mX_j, \mY) + \log(\rho_m/\rho_1)
	\right]}{
	\ds \sum_{\sl=1}^M \exp\left[ 
	\tfrac{1}{2}\lambda_{j\sl}(\mX_j, \mY) + \log(\rho_\sl/\rho_1)
	\right]
}
\\
& \ds = 2\sum_{m=2}^M \sum_{j \in \sO_k} \frac{  	
	\exp\left[ 
	\tfrac{1}{2} \widetilde{\lambda}_{jm}(\vx_j, \mY)
	+ \log(\rho_\sl/\rho_1)
	\right]}{
	 \sum_{\sl=1}^M \exp\left[ 
	\tfrac{1}{2}\widetilde{\lambda}_{j\sl}(\vx_j, \mY)
	+ \log(\rho_\sl/\rho_1)
	\right]
}

\end{array} 
$$

\noindent 
where 
$\widetilde{\lambda}_{jm}(\vx_j, \mY) = \lambda_{jm}(\vx_j, \mY) - \lambda_{j m_j}(\vx_j, \mY)$.

In the Supplementary material we show that $\rho_1,\ldots,\rho_M$ can be chosen to
achieve any desired penalty. Suppose that
$ 
\ds \log(\rho_m/\rho_1) = C \nu_m
$
 where $C$ is a tuning parameters. Then
$$
\tfrac{1}{2}\lambda_{jm}(\mX_j, \mY) - C\nu_m   = 
\mbox{IC}_{j1} - \mbox{IC}_{jm}
$$

\noindent where
$$
\ds \mbox{IC}_{jm} = - 2p_{jm}(\vx_j| \mY;\widehat{\vtheta}_{jm}) 
+ C  d_m 
$$

\noindent which can be thought of as a flexible information criteria. Note that
\begin{enumerate}
	\item $C = \log(n)$ corresponds to the Bayesian Information Criterion (BIC) \citep{Ref:BIC1978};
	
	\item $C = 2$ corresponds to the Akaike Information Criterion (AIC)  \citep{Akaike1974}; and
	
	\item $C = \log(n) + 2\log(p)$ corresponds to the Extended Bayesian Information Criterion (EBIC) \citep{EBIC}.
	 
\end{enumerate}

Let
$\widetilde{E}$ be an approximation of $E$ 
where the $\vx_j$ are replaced
with random draws under the data generating distribution
so that the $\widetilde{\lambda}_{jm}$'s appearing in the above expression for $E_\sO$
are asymptotically distributed as chi-squared random
variables with degrees of freedom $\widetilde{\nu}_m = \nu_{m} - \nu_{m_j}$.  
Using this chi-squared approximation of the LRT using the EBIC penalty above
we show that
$\widetilde{E} = o_p(1)$
provided $\log(p)/n \to 0$. Additional details and the proof of this
claim can be found in Appendix A.




The EBIC, unlike the BIC \citep{Ref:BIC1978}, rejects the assumption that the prior probabilities are uniform over all the possible models to be considered. Rather, the prior probability under the EBIC is inversely proportional to the size of the model class, reducing the number of features selected. 
The EBIC has been shown in   \cite{EBIC}   to work well in many contexts, such as high dimensional generalised linear models. However, we note if this algorithm is used in the scenario when $ n>p$, or if the true model space is not sparse, the EBIC may no longer be consistent. Further, flexibility can be provided through trading off FDR for positive selection rate (PSR) in order to yield improved predictive results.  

As such, we propose two penalties available to be used in this algorithm:

\begin{enumerate}
	\item A weaker penalty, in the form of the BIC \citep{Ref:BIC1978}, especially useful if used when $n > p$, or if improved PSR is preferred to low FDR \citep{EBIC}. In this case $\rho_1 \propto 1$; and,
	
	\item a stronger (default) penalty in the form of the EBIC.
\end{enumerate}

\section{Numerical Results} \label{numerical}

We will now assess the performance of our multiDA method, using both simulated and publicly available data to demonstrate ability in feature selection and in prediction under many different assumptions.

In Section \ref{multiDAcode}, we will describe the syntax
for our \code{R} package \code{multiDA}.
In Section \ref{competing} we
discuss competing methods to our own, and in the following Sections \ref{sims_FS}-\ref{sims_pred}, we will compare our multiDA algorithm with such methods with simulated data.
Lastly, Section \ref{benchmark}
compares each method publicly available data.
 
All of the following results were obtained in the \code{R} version 3.4.3 \citep{Rref} and all figures were developed throughout this paper using the \code{R} package \code{ggplot2} \citep{Ref:ggplot2}. Multicore comparisons were run on a 64 bit Windows 10 Intel  i7-7700HQ quad core CPU at up to 3.8GHz with 16GB of RAM.

\subsection{Using multiDA via the \code{R} package \code{multiDA}}\label{multiDAcode}

The syntax for \code{multiDA} is relatively straightforward:

\code{multiDA(mX,vy,penalty, equal.var,set.options)}

where the arguments of the \code{multiDA} function are explained below.

\begin{itemize}
	\item \code{mX} - data matrix with $n$ rows and $p$ columns.
	\item \code{vy} - a vector of length $n$ class labels, which can be a numerical or factor input.
	\item \code{penalty} - options used to define the penalty which will be used in the multiDA algorithm, as discussed in Section \ref{theo}, consisting of the options \code{EBIC} (default) and \code{BIC}.
	\item \code{equal.var} - the choice to run a multiLDA (\code{equal.var=TRUE}, default) or multiQDA (\code{equal.var=FALSE}) algorithm.
	\item \code{set.options} - the matrix partition to be used as described in Section \ref{hypothesis_test}. Options include \code{exhaustive} (default), \code{onevsrest}, \code{onevsall}, \code{ordinal}, and \code{sUser}, where \code{sUser} is a partition matrix provided by the user.
\end{itemize}

To predict new class labels, a generic S3 \code{predict(object,newdata)} command can be called with \code{\$vypred} returning predicted class labels and \code{\$probabilities} returning a matrix of class probabilities for each sample. An example finding a re-substitution error rate is provided below, using the \code{SRBCT} dataset as described in Section \ref{benchmark}.
 
\code{vy = SRBCT\$vy} \\
\code{mX = SRBCT\$mX} \\
\code{res = multiDA(mX, vy, penalty``EBIC", equal.var=T, set.options=``exhaustive") } \\
\code{vals = predict(res, newdata=mX)\$vy.pred}  \\
\code{rser = sum(vals!=vy)/length(vy)}

\subsection{Competing Methods}\label{competing}

We note that there is a long list of machine learning (ML) algorithms in the literature and as such we focus on comparison with representative algorithms from different broad classes of ML approaches. These are listed in Table \ref{tab:methods} below.

\begin{table}[ht!] \scriptsize  
	\centering
 	
	\begin{tabular}{ | m{5cm} | l |l|}
		\hline
		\textbf{Method} & \textbf{Paper} & \code{R} \textbf{Implementation}  \\ \hline
		DLDA/DQDA	    & \cite{dudoit2002} & \code{sparsediscrim} - \cite{Ref:sparsepackage} \\ \hline
		Penalized LDA   & \cite{Ref:penLDA}  & \code{penalizedLDA} - \cite{Ref:penLDApackage} \\ \hline
		Nearest Shrunken Centroids & \cite{Ref:tibshirani2003} & \code{pamr} - \cite{Ref:NSCpackage} \\ \hline
		Random Forest & \cite{Ref:Breiman2001} & \code{randomForest} - \cite{RFpackage} \\ \hline
		Support Vector Machine  (SVM) &  \cite{Ref:Cortes1995} & \code{e1071} - \cite{svmpackage} \\ \hline
		Multinomial logistic regression with LASSO regularization & \cite{Ref:lasso1996} & \code{glmnet} - \cite{glmnet} \\ \hline
		K nearest neighbours classifier ($K$=1) & \cite{Ref:KNNPaper} & \code{class} - \cite{class} \\
		
		\hline
	\end{tabular}
 	
	\caption {ML methods used in our comparisons} \label{tab:methods} 
\end{table}

Supplementary material contains the \code{R} code used for complete transparency.
 
\subsection{A simulation study - feature selection}\label{sims_FS}

In this section we assess the ability of multiDA to select correct informative features.
In particular we will empirically verify the theory described in Section
\ref{theo}.

We consider sample sizes $n=50$ to $n=500$ in increments of $50$, $p= 500$, $1000$, $5000$, $10000$, $20000$, and $K \in \{ 2, 3, 4, 5\}$, with the samples being equally distributed among the $K$ classes in each simulation. Note, we only run the simulation for when $p>n$, as we are purely interested in results in high dimensional space. Further, we consider a sparse feature space, such that only $10\%$ of the features are discriminative, with $s_k$ the set of such features determined to be discriminative.

In each simulation setting, data is generated as follows: For $j \in s_k$, sample from the space of non-null partitions of $\mS$, e.g., from a total of $M - 1 = 15$ non-null hypotheses for $K=4$. Simulate $\vx_{jmg} \sim \mathcal{N}(\mu_{jmg}, 1)$ where $g$ represents the group index of the normal distribution for the partition $m$ for feature $j$, and such that the mean shift for each differing normal distribution is $2$. If $j \notin s_k$, $\vx_i \sim \mathcal{N}(0,1)$.

\begin{figure}[ht!]
	\begin{center}
	\includegraphics[width=0.8\textwidth]{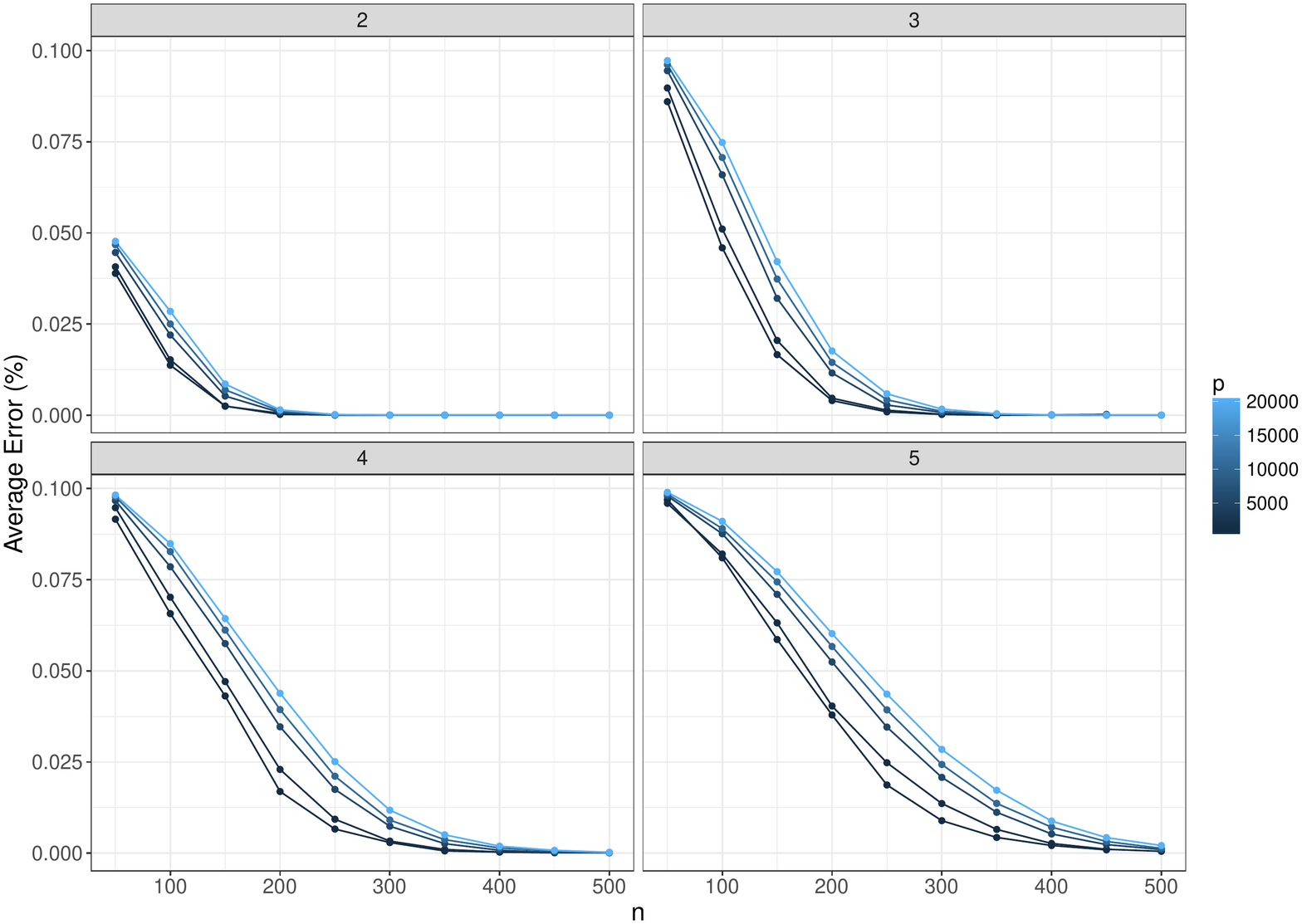}
\end{center}
	\caption{Simulation results - feature selection. Results have been faceted by values of $K$, with increasing $n$ on the x-axis and a separate curve for each $p$. }
	\label{fig:Sim}
\end{figure}

We simulate data using the above 20 times and the average
total proportion incorrectly selected features (given by $E/M$) 
over these 20 replications is displayed in Figure \ref{fig:Sim}. 
It is clear that as $n \rightarrow \infty$, the error rates asymptotically converge to 0, regardless of the values of $p$ and $K$. 
However, in the case when $ p\gg n$ and $n$ small, the error increases (as to be expected) but is no bigger than $10\%$ when $K=5, p=20000$, and $n=50$.

\subsection{A simulation study - prediction}\label{sims_pred}

In this study we assess the predictive performance of the multiDA algorithm with simulated data. Four simulations were considered, with two simulating under the assumption of independent features, and the other two simulating more realistic data with multivariate normal data generated using sparse covariance matrices. In all simulations, we consider predicting the four classes, with our data matrices of dimension $n=100, p=20000$. As in Section \ref{sims_FS}, we consider a highly sparse set of discriminative features, utilising the sets $s_k$ as defined previously. Finally, as before, samples are spread equally among the four classes. The specifics of simulation settings are detailed below.

\textbf{Simulations with Independent Features}

\begin{enumerate}
	\item \textit{Mean shift with independent features, equal group variances} \\
	For $j \in s_k$, sample from the space of non-null partitions of $\mS$ (total of 15 non null for $K=4$). Simulate $\vx_{jmg} \sim \mathcal{N}(\mu_{jmg}, 1)$ where $g$ represents the group index of the normal distribution for the partition $m$ for feature $j$, such that the mean shift for each differing normal distribution is $0.5$. If $j \notin s_k$, $\vx_i \sim \mathcal{N}(0,1)$.
	
	\item \textit{Mean shift with independent features, unequal group variances}\\
	Same as 1), however, the group variances are allowed to change within each partition, such that for $j \in s_k$, simulate $\vx_{jg} \sim \mathcal{N}(\mu_{jmg}, \sigma^2_{jmg})$, where $\mu_{jg}$ is defined as in 1), and such that the scale in variance for each normal distribution is $1$.
	
\end{enumerate}

\textbf{Simulations with Dependent Features}

For the simulations below, define the set $s_k$ such that $s_1 = [1,500], s_2 = [501,1000], s_3=[1001, 1500]$, and $s_4= [1501,2000]$.
\begin{enumerate}
	\setcounter{enumi}{2}
	\item \textit{Dependent features, equal group covariances}\\
	For $i \in \vy_k, j \in s_k$, $\vx \sim \mathcal{N}(0.5, \mSigma)$ and $\vx_i \sim \mathcal{N}(0, \mSigma)$ for $j \notin s_k$. $\mSigma$ is generated as a sparse covariance matrix, such that blocks of size $2000 \times 2000$ have high correlation.
	\item \textit{Dependent features, unequal group covariances}\\
	Identical to (3), except that for $i \in \vy_k, j \in s_k$, $\vx \sim \mathcal{N}(0.5, \mSigma_k)$, as in this case we consider the scenario in which group covariances differ.

\end{enumerate}

The $\mSigma$ and $\mSigma_k$ covariance matrices are constructed as follows. Let $\mB_\ell$ be a
$b \times b$ matrix where the diagonal entries and
$\Delta$ percent of the off diagonal entries are generated using
independent normally distributed random variables. Then the
matrices $\mSigma$ and $\mSigma_k$ are constructed by
taking 10 matrices $\mB_1,\ldots,\mB_{10}$ with $b=2000$ with
$\Delta = 0.25$, forming the block diagonal matrix
consisting of the blocks $\mB_1^T\mB_1,\ldots, \mB_{10}^T\mB_{10}$
and then permuting the rows and columns of the resulting matrix.

\begin{figure}[ht!]
 
	\begin{center} 
	\includegraphics[width=\textwidth]{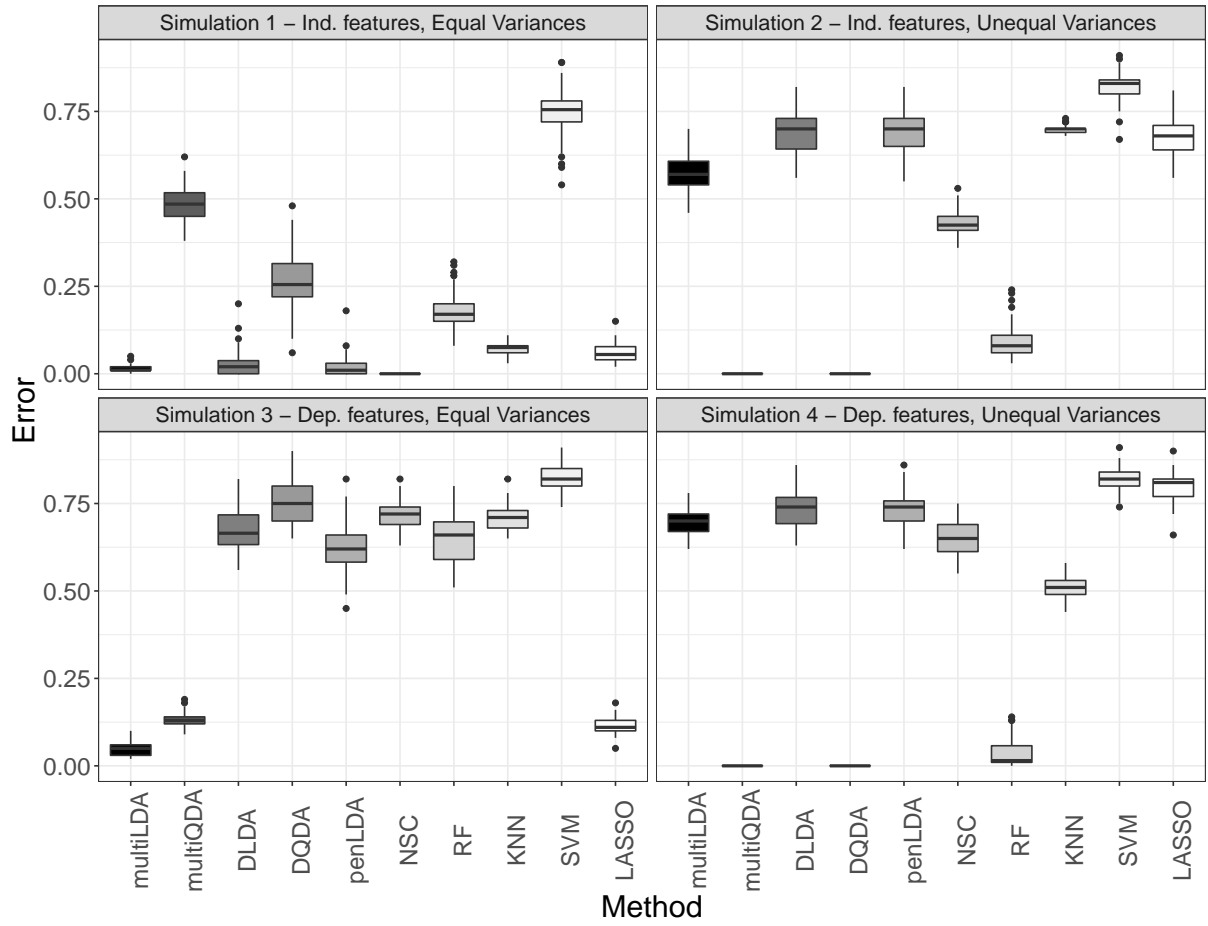}
\end{center}
	\caption{50 trial, 5 fold cross validation errors for simulated data as described in \ref{sims_pred}. Independent feature simulations are on the top facets, whilst dependent feature results are on the bottom.}
	\label{fig:Sim2}
\end{figure}

50 x 5 fold cross validation results are shown in Figure \ref{fig:Sim2}.

\subsection{Performance on publicly available datasets for benchmark data comparison} \label{benchmark}

We run our multiDA algorithm, among others, on three separate publicly available data sets, for benchmark data comparison. For all three datasets, we run a 50 x 5 fold cross validation, as well as report the algorithmic run times for running the algorithm once on the full data set.

\newpage 

\textbf{TCGA Breast Cancer data}

\begin{figure}[ht!]
	\begin{center} 
	\includegraphics[width=0.8\textwidth]{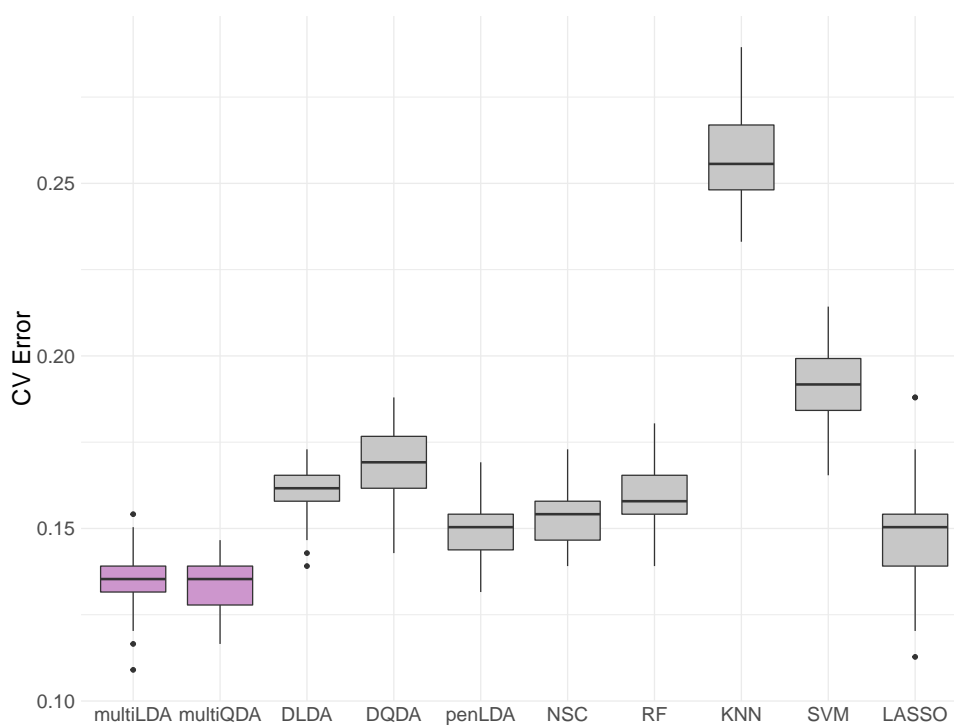}
\end{center}
	\caption{50 x 5 fold cross validation TCGA Breast Cancer results. multiDA results in violet with competing methods in grey.}
	\label{fig:TCGA}
\end{figure}

Microarray data from the Cancer Genome Atlas (TCGA) was used to classify 5 different subtypes of breast cancer, namely ``Basal", ``HER2", ``Luminal A", and ``Luminal B", as well as distinguish between healthy tissue, with the data consisting of 266 samples and a feature set of size 15803. As the breast cancer subtypes were defined intrinsically based on PAM50 gene expression levels, these genes were removed first in order to see if other complementary genes could be used in order to predict cancer subtype. Cross validation results are shown in Figure \ref{fig:TCGA}. Results for the multiDA algorithms were achieved using default settings for our algorithm.

\newpage 

\textbf{Small Round Blue Cell Tumors (SRBCT) data}

\begin{figure}[ht!]
	\begin{center} 
	\includegraphics[width=0.8\textwidth]{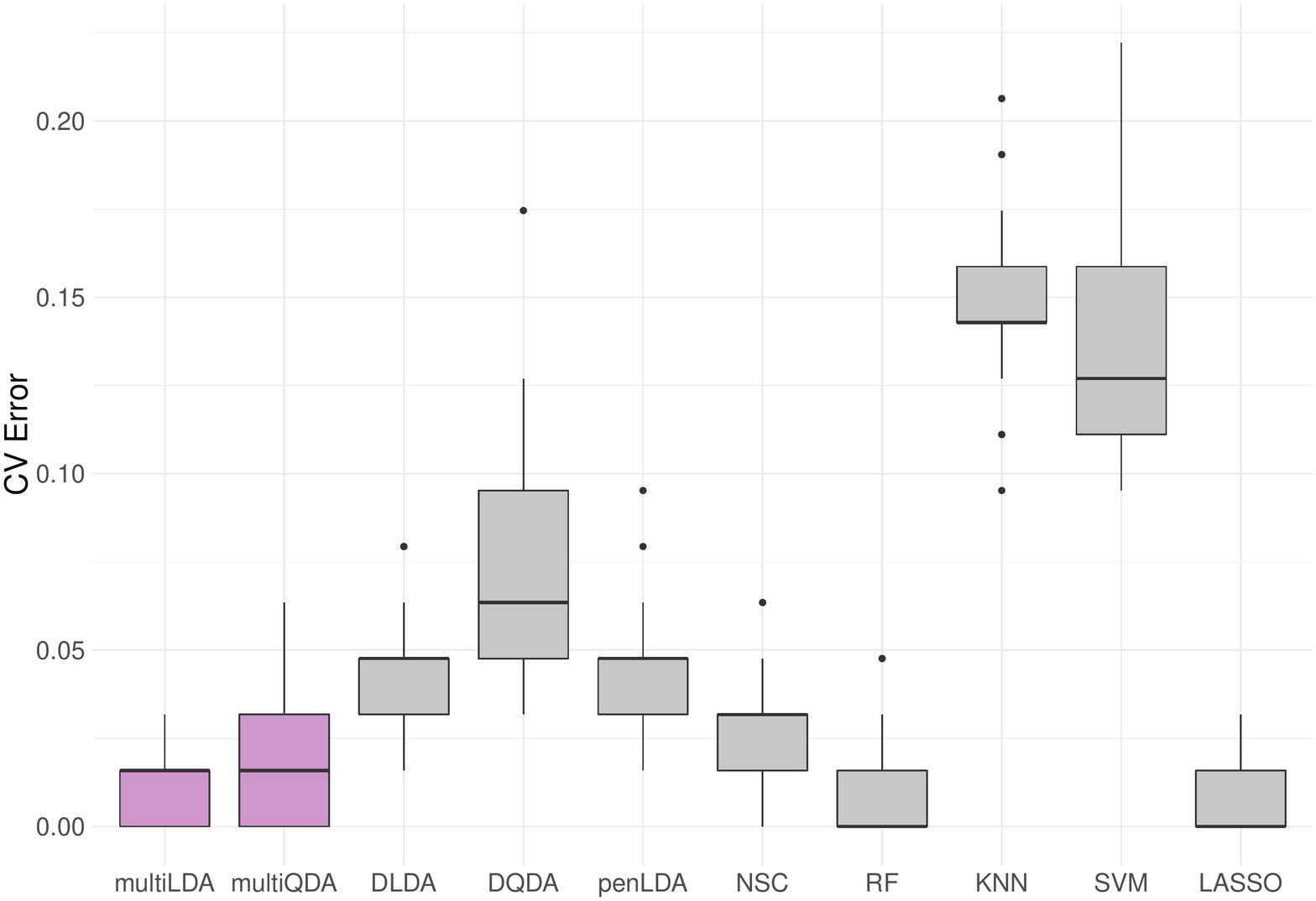}
	\end{center}
	\caption{50 x 5 fold cross validation SRBCT results. multiDA results in violet with competing methods in grey.}
	\label{fig:SRBCT}
\end{figure}

The SRBCT dataset \citep{Khan2001} looks at classifying 4 classes of different childhood tumours sharing similar visual features during routine histology. These classes include Ewing's family of tumours (EWS), neuroblastoma (NB), Burkitt's lymphoma (BL), and rhabdomyosarcoma (RMS). Data was collected from 83 cDNA microarrays, with 1586 features present after filtering for genes with zero median absolute deviation.  Cross validation results are shown in Figure \ref{fig:SRBCT}. Results for the multiDA algorithms were achieved using default settings for our algorithm.

\textbf{Melanoma data}

The melanoma dataset has been analysed by \citep{Ref:Mann2013} as a binary classification problem. The dataset consist of $n = 98$ samples of Affymetrix arrays with $p = 23901$ genes measured on each array.
The response variable is the patient's prognosis, which is compromised of three classes - good, middle, and poor prognosis groups.  Class labels have been defined as follows, using the status of the patient and their survival times:
\begin{align*}
y_i =
\begin{cases} 
\hfill 1    \hfill & \text{ if $t_i < 2$ and Dead Melanoma \quad \quad \quad \hspace{5mm} ``Poor" prognosis.} \\
\hfill 2 \hfill & \text{ if $2 < t_i < 6$ and Dead Melanoma \quad \quad \hspace{2mm} ``Middle" prognosis.} \\
\hfill 3 \hfill & \text{ if $t_i > 6$ and Alive No Sign of Relapse \quad ``Good" prognosis.} \\
\end{cases}
\end{align*}

\begin{figure}[ht!]
	\begin{center} 
	\includegraphics[width=0.8\textwidth]{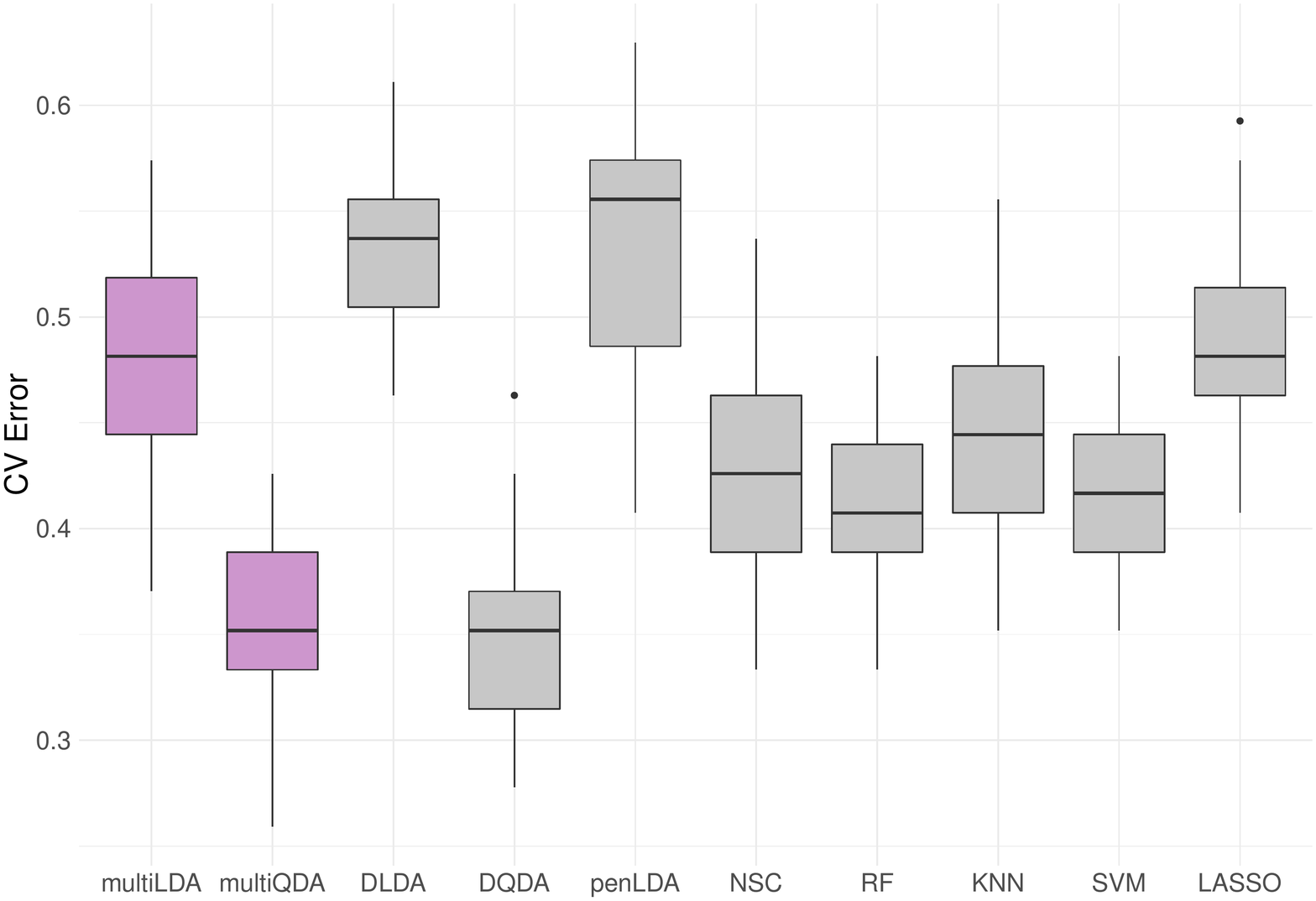}
\end{center}	
	
	\caption{50 x 5 fold cross validation Melanoma results. multiDA results in violet with competing methods in grey.}
	\label{fig:Melanoma}
\end{figure}

Samples that do not meet the criteria above are excluded, leaving $n=54$ samples for analysis. Further, we implemented a filtering step that excludes under-expressed genes (all three class medians below 7), resulting in $p=12404$ genes to be used for analysis. Cross validation results are shown in Figure \ref{fig:Melanoma}. 

Results for the multiDA algorithms were achieved using default settings except for the penalty - in which the BIC setting was used. The use of a weaker penalty is justified through an extreme trade-off between PSR and FDR. Investigation into features selected using the EBIC penalty revealed that those deemed significant by multiQDA to be driven by extreme outliers - due to the strong penalty ensuring that only those with a maximal LRT are deemed discriminative, greatly reducing the PSR. As such, we have selected the BIC penalty, acknowledging the likely increase in FDR, however improving predictive performance by increasing PSR (strong predictive performance of DQDA indicates importance of PSR).

 \subsection{Timings} 
 
While we recognise that run time of each algorithm depends on various factors involving hardware and  implementation chosen, we provide the timings in
Table \ref{tab:times} below as indicative in the context of the details outlined
in Section  \ref{numerical}. These times suggest that
multiLDA and multiQDA both run in a reasonable amount of
time in comparison to the competing methods considered in this paper.
 
\begin{table}[ht!]\scriptsize
	\centering
	
	\begin{tabular}{ | l | l |l|l|l|l|l|l|l|l|l |l }
		\hline
		& \textbf{multiLDA} & \textbf{multiQDA} & \textbf{DLDA} & \textbf{DQDA} & \textbf{penLDA}  & \textbf{NSC} & \textbf{RF}  & \textbf{KNN} & \textbf{SVM} & \textbf{LASSO} \\ \hline
		TCGA & 30.76 & 71.42 & 1.47 & 3.95 & 199.78 & 11.31 & 63.94  & 6.73 & 10.47 & 37.27 \\          
		SRBCT & 0.24 & 0.55 & 0.11 & 0.22 & 17.92 & 0.76 & 1.24 & 1.22 & 0.22 & 1.55  \\
		Melanoma            & 0.42 & 0.88 & 0.77 & 1.14 & 8.17 & 5.86 & 7.73 & 0.16 & 1.39 & 6.84 \\ 
		\hline
	\end{tabular}

\caption {Timings for a single run in (s) for benchmark datasets} \label{tab:times} 
\end{table}

\subsection{Discussion of results}

Through simulation studies and applications on publicly available data, we are able to analyse the strengths of our multiDA classifier. Our simulation study examining the feature selection component of multiDA demonstrates the ability of multiDA to accurately select the correct feature set as the sample size increases, confirming our theoretical considerations. As the number of classes increases, the sample size needed to select the correct features increase, however this is to be expected.  

Examining predictive strength through both simulations and benchmark data analysis, it is clear that the multiLDA algorithm excels when group variances are assumed to be equal (Simulations 1 and 3), and performs poorly when this assumption is not met (Simulations 2 and 4, Melanoma microarray analysis). The converse is true for multiQDA, which also shows improvements to all LDA based methods in the unequal variances cases as well as out performing plain DQDA in Simulation 2 and also showing strengths in the Melanoma example. In the independent features case for our simulation studies, the other DA methods (penLDA, NSC, DLDA, DQDA) do well as expected, however their performance is much worse in Simulation 2 (as expected). SVM has poor performance overall in both simulated and benchmark data examples, with KNN, LASSO, and RandomForest (RF) performing better in some simulations as compared to others.

It is also clear that the multiDA classifier can perform well when assumptions of independence of features is not met. This is demonstrated in Simulations 3 and 4 when we consider data with a non diagonalised covariance structure, and also in the benchmark data analysis examples, where in gene expression data we expect some degree of correlation between the features. 

In all examples shown with the exception of Simulation 1, the multiDA classifiers have performed as well or better than all DA-like methods, consistently outperformed methods such as KNN and SVM, and have produced competitive results with Random Forest and the LASSO on multiple datasets. 

\section{Conclusion and future work}\label{conclusion}

We have introduced the multiDA classifier in order to provide an effective alternative to discriminant analysis in high dimensional data. We have utilised a multiple hypothesis testing paradigm in order to select relevant features for our algorithm, utilising latent variables and penalised likelihood ratio tests to do so. Not only is the procedure intuitive and fast, we have shown the feature selection process is consistent given appropriate penalties. Further, as shown in simulation and benchmark data analysis studies, our classifier yields prediction results that are competitive with not only other discriminant analysis methods, but also other non linear machine learning methods as well. We believe that the multiDA classifier is a useful tool for any analyst wanting fast and accurate classification results for high dimensional Gaussian data. 

Future work can be done to extend the scope of the distributions the multiDA classifier can effectively model. For example, the datasets provided in Section \ref{benchmark} were microarray data. By extending the multiDA classifier to handle negative binomial or Poisson data, datasets such a RNA-Seq could potentially be well predicted using the core ideas of multiDA. 

\section*{Acknowledgements} 

We are grateful for the data provided to us, processed as described in \citep{Network2012} by Kim-Anh Le Cao (University of Melbourne). This project benefited from conversations with Samuel M\"uller.
This research was partially supported by an Australian Postgraduate Award (Romanes), an
Australian Research Council
Early Career Award DE130101670 (Ormerod) and Australian Research Council Discovery Project grant
DP170100654 (Yang, Ormerod); Australia NHMRC Career Developmental Fellowship APP1111338 (Yang).

\newpage 

\bibliography{BibMCDA}

\section*{Appendix A - Theory}

Let $X_{ij} \sim p_{jm_j}(\, \cdot \, ; \vtheta_{0jm})$
for some true parameter vector $\vtheta_{0jm} \in \bR^{d_{m}}$.
Define the log-likelihood for variable $j$ and
hypothesis $m$ as
$\sl_{jm}(\vtheta) 
=  \sum_{i=1}^n \log p_{jm}(X_{ij};\vtheta_{jm})$
with corresponding MLE
and ``pseudo-true'' value of $\vtheta_{jm}$ as
$$
\ds \widehat{\vtheta}_{jm} = \mbox{arg}\max_{\vtheta_{jm}} \left\{ \sl_{jm}(\vtheta_{jm}) \right\}
\qquad \mbox{and} \qquad 
\ds \vtheta_{jm}^* = \mbox{arg}\max_{\vtheta_{jm}} \left\{ \bE\left( n^{-1}\sl_{jm}(\vtheta_{jm}) \right) \right\},
$$

\noindent respectively. We will assume conditions on the likelihood and
parameter space such that
$\bE\left[ n^{-1}\sl_{jm}(\vtheta_{jm}^*) \right] \to \sl_{jm}^*$ 
for $1\le j\le p$, $1\le m\le M$. Using the theory summarised in
\cite{OrmerodEtal2017} based on \cite{vuong89} and \cite{vaart98}
we have two main cases to consider.
\begin{itemize}
	\item {\bf[Underfitting case]} -- Suppose $\sl_{jm_j}^* > \sl_{jm}^*$ for some $m\ne m_j$. Then
	$$
	\tfrac{1}{2n}\left[ \lambda_{jm_j}(\mX_j) - \lambda_{jm}(\mX_j) \right] \stackrel{P}{\to} \sl_{jm_j}^* - \sl_{jm}^* = \Delta_{jm} > 0
	$$
	
	\noindent and so $(1/2)\lambda_{jm}(\mX_j) = n[\Delta_{jm} + o_p(1)]$.
	
	\item {\bf[Overfitting case]} -- Suppose $\sl_{jm}^* = \sl_{jm_j}^*$ for some $m\ne m_j$ and let $\nu_{m} = d_{m} - d_{m_j}$. Then
	$$
	\lambda_{jm}(\mX_j) - \lambda_{jm_j}(\mX_j) \stackrel{D}{\to} \chi_{\widetilde{\nu}_{jm}}^2  
	$$
\end{itemize}

\noindent 

\noindent The following lemma will be useful later.

\noindent 
{\bf Lemma 1 \citep{Gasull2015}:} 
If $X_j$, $j=1,\ldots,p$, are independent $\chi_{\nu}^2$ 
random variables, and
$\ds M_p = \max_{1\le j\le p} \{ X_{j} \}$,
then
$$
\tfrac{1}{2}M_p - \left[
\log(p) + (\nu/2 - 1)\log\log(p) - \log\Gamma(\nu/2)
\right] \stackrel{D}{\to} G,
$$	

\noindent as $p\to \infty$ where $G$ is a Gumbel distributed random variable.



Define
$\sJ_m = \{ j \colon \gamma_{0jm} = 1 \}$ and
$\sT_m = \{ j \colon \sl_{jm}^* = \sl_{jm_j}^* \}$.
Here $\sJ_k$ is the set of true variables over the $k$th set
of hypotheses, and $\sT_m$ is the union of over-fitting and true models over the $k$th set of hypotheses.
We define and decompose the the error as
$$
\begin{array}{rl}
E 
%
& \ds 
= 
\sum_{j=1}^p 1 - \widehat{\gamma}_{jm_j}(\mX_j)
+ 
\sum_{j=1}^p \sum_{m \ne m_j} \widehat{\gamma}_{jm}(\mX_j)


\\
& \ds 
=  
2\sum_{j=1}^p \sum_{m \ne m_j} \widehat{\gamma}_{jm}(\mX_j)
=  
2\sum_{m=1}^M \sum_{j \notin \sJ_m} \widehat{\gamma}_{jm}(\mX_j)
\\ 
& \ds =
\underbrace{
	2\sum_{m=2}^M \sum_{j \in \sO_m} \widehat{\gamma}_{jm}(\mX_j)
}_{
	\mbox{Overfitting models}
} 
+ 
\underbrace{
	2\sum_{m=1}^M \sum_{j \in \sU_m} \widehat{\gamma}_{jm}(\mX_j)
}_{
	\mbox{Underfitting models}
} 
\\
& \ds \stackrel{\triangle}{=}
E_\sO + E_\sU,

\end{array} 
$$

\noindent where $\sO_m = \sJ_m^c \cap \sT_m$, and $\sU_m = \sJ_m^c \cap \sT_m^c$.

\noindent Note that for $E_\sO$ the index $m$ is summation
does not include $m=1$ since the null model cannot be an overfitting model. Next, we consider $E_\sO$ where the true model is used as the null hypothesis and rewrite $E_\sO$ as
$$
\begin{array}{rl}
E_\sO


& \ds = 2\sum_{m=2}^M \sum_{j \in \sO_m} \frac{  	
	\exp\left[ 
	\tfrac{1}{2} \widetilde{\lambda}_{jm}(\mX_j)
	- \widetilde{\nu}_k \left\{ \log(n)
	+ 2 \log(p) \right\}
	\right]}{
	 \sum_{\sl=1}^M \exp\left[ 
	\tfrac{1}{2}\widetilde{\lambda}_{j\sl}(\mX_j)
	- \widetilde{\nu}_\sl \left\{ \log(n)
	+ 2 \log(p) \right\}
	\right]
}

\end{array} 
$$

\noindent
where $\widetilde{\lambda}_{jm}(\mX_j) = \lambda_{jm}(\mX_j) - \lambda_{jm_j}(\mX_j)$.
Using a chi-square approximation over the set of over-fitting models in place of LRT statistics with
$U_{jm} \stackrel{\mbox{\scriptsize iid}}{\sim} \chi_{\widetilde{\nu}_m}^2$
we obtain an
approximation $\widetilde{E}_\sO$ of $E_\sO$    given by
$$
\begin{array}{rl}
\ds \widetilde{E}_\sO
& \ds = 2\sum_{m=1}^M \sum_{j \in \sJ_m^c \cap \sT_m} \frac{\ds   	
	\exp\left[ 
	\tfrac{1}{2} U_{jm} 
	- \widetilde{\nu}_m \left\{ \log(n)
	+ 2 \log(p) \right\}
	\right]}{
	 \sum_{\sl=1}^M \exp\left[ 
	\tfrac{1}{2}\lambda_{j\sl}(\mX_j)
- \widetilde{\nu}_\sl \left\{ \log(n)
+ 2 \log(p) \right\}
	\right]
}

\\
& \ds \le  2\sum_{m=1}^M \sum_{j \in \sO_m} 
\exp\left[ 
\tfrac{1}{2} U_{jm} - \widetilde{\nu}_m \left\{ \log(n)
+ 2 \log(p) \right\}
\right]

\\
& \ds \le   2\sum_{m=1}^M p_{1m}
\exp\left[ 
 \max_{j \in \sO_m} \tfrac{1}{2} Z_{jm}   - \widetilde{\nu}_k \left\{ \log(n)
+ 2 \log(p) \right\}
\right]

\\
& \ds \to  \sum_{m=1}^M  \left( \frac{p_{1m}^2}{p^{2\widetilde{\nu}_m} \log(p_{1m})} \right) \left( \frac{\log(p_{1m})^{1/2}}{n} \right)^{\widetilde{\nu}_m} \frac{2\exp(
	G_m)}{\Gamma(\widetilde{\nu}_m/2)}

\end{array} 
$$

\noindent where
$p_{1k} = |\sO_m|$, the last line
follows from Lemma 1 with $G_1,\ldots,G_m$ being
independent Gumbel distributions. Note that $E_\sO = o_p(1)$ provided
$\log(p)/n \to 0$.
Similarly, for $E_\sU$ we have
$$
\begin{array}{rl}
\ds E_\sU
& \ds \le 2\sum_{m=1}^M \sum_{j \in \sU_m} 
\exp\left[ 
\tfrac{1}{2}\widetilde{\lambda}_{jm}(\mX_j) - 
\widetilde{\nu}_m \{ \log(n)
+ 2\log(p) \}
\right]
\\
& \ds = 2\sum_{m=1}^M \sum_{j \in \sU_m} 
\exp\left[ 
- \tfrac{1}{2} n\{ \widetilde{\Delta}_{jm} + o_p(1) \} - 
\widetilde{\nu}_m \{ \log(n)
+ 2\log(p) \}
\right]
\\
& \ds \le 2\sum_{m=1}^M \frac{ p_{0m}}{p^{
		2\widetilde{\nu}_m}} 
\exp\left[ 
- \frac{1}{2} n \left\{ \min_{j \in \sU_m}  \widetilde{\Delta}_{jm} \right\} - \widetilde{\nu}_m\log(n)
\right] + \mbox{smaller terms}
\\
& \ds = o_p(1)
\end{array} 
$$

\noindent where $\widetilde{\Delta}_{jm} = \Delta_{jm_j} -\Delta_{jm} > 0$,
the above $\widetilde{\nu}_m$ may be positive or negative,  and $p_{0m} = |\sU_m|$.
The only potentially problematic term occurs for when
$m=1$ since $\nu_1 = 0$. For this case $E_\sU = o_p(1)$ provided
$$
p_{01}\exp\left[ 
- \frac{1}{2} n \left\{ \min_{j \in \sU_1}  \widetilde{\Delta}_{j1} \right\}
\right] = o(1).
$$

\bigskip
\noindent 
Which is true provided $\log(p)/n\to 0$. Hence, $ \widetilde{E}_\sO + E_\sU = o_p(1)$.

\end{document}